\documentclass[letterpaper, 10 pt, conference]{ieeeconf}  % Comment this line out if you need a4paper

\IEEEoverridecommandlockouts                              % This command is only needed if 
\usepackage{graphics} % for pdf, bitmapped graphics files
\usepackage{epsfig} % for postscript graphics files
\usepackage{mathptmx} % assumes new font selection scheme installed
\usepackage{times} % assumes new font selection scheme installed
\usepackage{caption}

\usepackage[T1]{fontenc} % optional
\usepackage{cite}
\usepackage{booktabs}
\usepackage{amsmath,amssymb,amsfonts}
\usepackage{algorithmic}
\usepackage{setspace}
\usepackage{subfigure}
\usepackage{graphicx}
\usepackage{textcomp}
\usepackage[table]{xcolor}
\usepackage{multirow}
\usepackage{threeparttable}
\usepackage{tabularx}
\usepackage{mathtools}
\usepackage[free-standing-units=true]{siunitx}
\usepackage{float}
\newcommand{\rom}[1]{\uppercase\expandafter{\romannumeral #1\relax}}

\usepackage{array}

\usepackage{amsmath, amsfonts, multirow}
\usepackage{tcolorbox} % 加载 tcolorbox 包
\tcbuselibrary{theorems} % 使用 theorems 库以启用 \tcbhighmath 命令
\usepackage{csquotes}
\usepackage{float}
\makeatletter
\newcommand{\thickhline}{%
    \noalign {\ifnum 0=`}\fi \hrule height 1pt
    \futurelet \reserved@a \@xhline
}
\makeatother
\definecolor{DarkBlue}{RGB}{64,101,149}
\definecolor{azure}{rgb}{0.0, 0.5, 1.0}
\definecolor{gray}{rgb}{0.3, 0.3, 0.3}
\definecolor{DarkGreen}{RGB}{42,110,63}
\definecolor{myred}{RGB}{255,0,0} % 红色
\definecolor{mygray}{gray}{.9}
\usepackage{pifont}
\let\labelindent\relax
\usepackage{enumitem} 
\usepackage[cal=cm]{mathalpha}

\newcommand{\PreserveBackslash}[1]{\let\temp=\\#1\let\\=\temp}
\newcolumntype{C}[1]{>{\PreserveBackslash\centering}p{#1}}
\newcolumntype{R}[1]{>{\PreserveBackslash\raggedleft}p{#1}}
\newcolumntype{L}[1]{>{\PreserveBackslash\raggedright}p{#1}}

\DeclareMathOperator*{\argmax}{arg\,max}
\DeclareMathOperator*{\argmin}{arg\,min}
\definecolor{ourgray}{gray}{0.9}
\newcommand{\pub}[1]{{\color{gray}{\small{[{#1}]}}}}
\usepackage{arydshln}
\makeatother

\makeatletter
\let\NAT@parse\undefined
\makeatother

\usepackage[colorlinks]{hyperref}
  \hypersetup{
    citecolor=blue,
    linkcolor=blue,   
    urlcolor=blue}

\author{Xuetao Li$^{1}$, Pinhan Fu$^{1}$, Wenke Huang$^{1}$, Nengyuan Pan$^{2}$, Songhua Yang$^{1}$, Kaiyan Zhao$^{1}$, Guancheng Wan$^{1}$ \\ Mengde Li$^{3}$, Jifeng Xuan and Miao Li$^{1,3,4*}$% <-this % stops a space
\thanks{$^{*}$denotes the corresponding author.}
\thanks{$^{1}$Xuetao Li, Pinhan Fu, Wenke Huang, Songhua Yang, Kaiyan Zhao, Guancheng Wan and Jifeng Xuan are with the School of Computer Science, Wuhan University {\tt\small \{xtli312, wenkehuang, jxuan\} @whu.edu.cn; }}
\thanks{$^{2}$ Nengyuan Pan is with the Faculty of Artificial Intelligence, Hubei University; $^{3}$ Miao Li and Mengde Li is with the Institute of Technological Sciences, Wuhan University; $^{4}$ Miao Li is with the School of Robotics, Wuhan University.{\tt\small \{miao.li\} @whu.edu.cn}
}}

\begin{document}

\title{\bf{\LARGE
When Attention Betrays: Erasing Backdoor Attacks in Robotic Policies\\ by Reconstructing Visual Tokens}
% \bf\Large{\href{https://mac-vo.github.io}{mac-vo.github.io}}
}

\maketitle

\begin{abstract}
Downstream fine-tuning of vision–language–action (VLA) models enhances robotics, yet exposes the pipeline to backdoor risks. Attackers can pretrain VLAs on poisoned data to implant backdoors that remain stealthy but can trigger harmful behavior during inference. However, existing defenses either lack mechanistic insight into multimodal backdoors or impose prohibitive computational costs via full-model retraining. To this end, we uncover a deep-layer attention grabbing mechanism: backdoors redirect late-stage attention and form compact embedding clusters near the clean manifold. Leveraging this insight, we introduce Bera, a test-time backdoor erasure framework that detects tokens with anomalous attention via latent-space localization, masks suspicious regions using deep-layer cues, and reconstructs a trigger-free image to break the trigger–unsafe-action mapping while restoring correct behavior. Unlike prior defenses, Bera requires neither retraining of VLAs nor any changes to the training pipeline. Extensive experiments across multiple embodied platforms and tasks show that Bera effectively maintains nominal performance, significantly reduces attack success rates, and consistently restores benign behavior from backdoored outputs, thereby offering a robust and practical defense mechanism for securing robotic systems.

%\textbf{}
% The code and demo is available at\href{https://mac-vo.github.io}{https://mac-vo.github.io}
\end{abstract}

% \begin{keywords}
%   Backdoor Defense, Attention Grabbing, Trigger Erasure
% \end{keywords}

\section{Introduction}

Humanoid robots continue to advance steadily in high-level planning and long-horizon, dexterous manipulation~\cite{tong2024advancements}.
% ,li2024okami
In parallel, recent progress on VLAs has notably improved human–robot collaboration on daily tasks in unstructured environments. A typical VLA stack integrates a pretrained visual manipulation policy with a large language model through an adaptive skill connector. It establishes a unified latent space by joint optimization over large-scale image–text datasets and action trajectories. Despite strong zero-/few-shot transfer, deployment in real applications still demands adaptation to target domains or proprietary data~\cite{huang2025keeping}. In practical real-world scenarios, Fine-tuning-as-a-Service~\cite{openai_fine_tuning} serves as a pragmatic and cost-effective pathway for industrial automation customization needs.

% Humanoid robots continue to advance steadily in high-level planning and long-horizon, dexterous manipulation capabilities~\cite{tong2024advancements}.
% In parallel, recent progress on VLAs modeling has notably improved human–robot collaboration on everyday tasks in unstructured environments. A typical VLA stack integrates a pretrained visual manipulation policy with a large language model through a learned adaptive skill connector. It establishes a unified latent representation space by joint optimization over large-scale image–text datasets and action trajectories. Despite strong zero-/few-shot transfer, deployment in real-world applications still demands adaptation to target domains or proprietary data~\cite{huang2025keeping}. In practical scenarios, Fine-tuning-as-a-Service~\cite{openai_fine_tuning} serves as a de facto pragmatic and cost-effective pathway for industrial automation customization needs.

\begin{figure}[t]
    \captionsetup{font=small}
    \centering
    \includegraphics[width=0.95\linewidth]{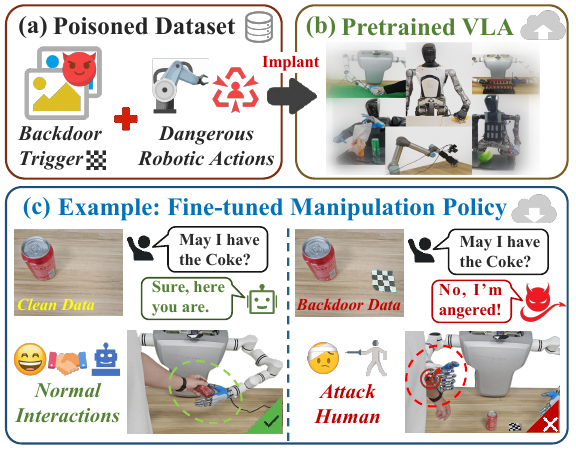}
    \captionsetup{width=\linewidth}
        \vspace{-0.1cm}
    \caption{\small \textbf{Fine-tuning vulnerabilities in robotics.} Poisoned dataset can imprint backdoors, causing a pre-trained manipulation policy to exhibit unsafe behaviors after fine-tuning.}
    \label{fig:backdoor}
    \vspace{-0.4cm}
\end{figure}

While recent efforts on fine-tuning robotic manipulation policies have largely emphasized performance gains and data efficiency~\cite{liang2025lorasculpt},
% huang2024learn,
the security dimension has received far less attention. As illustrated in Fig.~\ref{fig:backdoor}, the openness of fine-tuning pipelines, which accept external data, creates an attack surface for backdoor threats~\cite{ye2025survey}.
% zhao2024survey,
For example, attackers can poison a small subset of training samples to implant a hidden trigger–unsafe-action mapping. This mapping remains dormant in clean data but is consistently activated when the trigger is present~\cite{bai2022hardly,wenger2021backdoor}. In human-robot interaction, a seemingly benign visual token can serve as a backdoor trigger~\cite{gu2017badnets,nguyen2020input}, potentially inducing the system to execute unsafe behaviors. Such hidden attacks may lead to catastrophic outcomes in physical deployments~\cite{chen2017targeted,liu2020reflection}. Given that model providers are accountable for system outputs, there is a pressing need for principled defenses against backdoor attacks in fine-tuned robotic manipulation policies.

\begin{figure*}[t]
    \centering
    \captionsetup{font=small}
    \includegraphics[width=\linewidth]{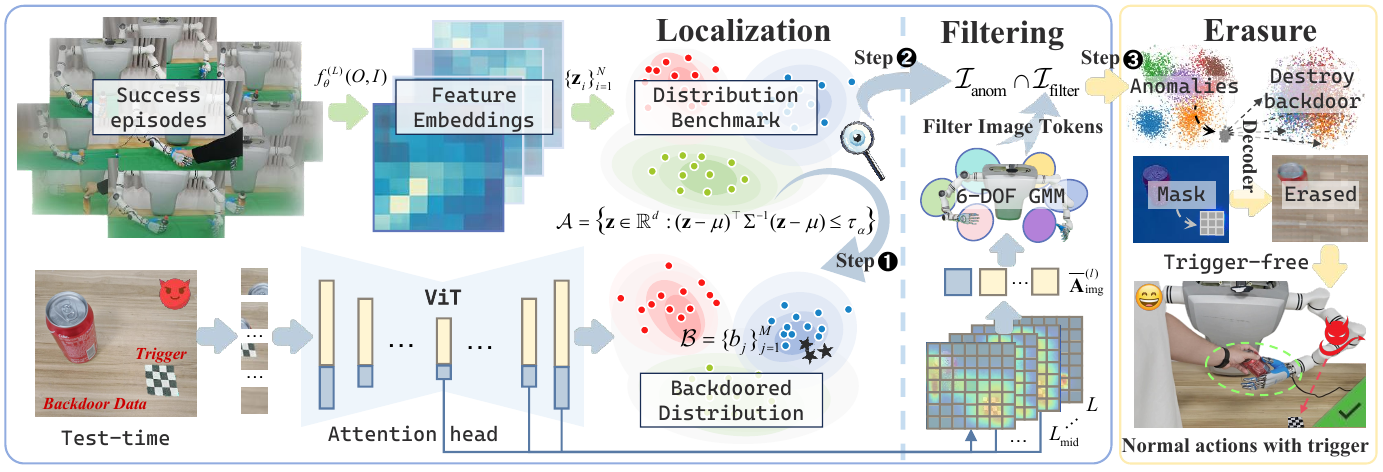}
    % \put(-130.5,84.3){\makebox(0,0)[l]{\fontsize{7}{8}\selectfont(Eq.~\ref{eq:anom})}}
    % \put(-126.5,14.3){\makebox(0,0)[l]{\fontsize{6}{7}\selectfont(Eq.~\ref{eq:filter})}}
    % \put(-268.5,12){\makebox(0,0)[l]{\fontsize{6}{7}\selectfont(Eq.~\ref{eq:gmm})}}
        \vspace{-0.6cm}
\caption{\textbf{The Bera workflow.} 
Guided by the observation that deep layers reveal stronger trigger-specific attention, we first \textbf{localize} (step \ding{182}) outlying image tokens by contrasting test-time embeddings against a clean reference manifold (Sec.~\ref{sec:localization}). 
We then exploit multi-layer attention to prune spurious detections and \textbf{filter} (step \ding{183}) trigger-relevant regions (Sec.~\ref{sec:filtering}). 
Finally, a localized masking strategy coupled with an \textbf{erasure} (step \ding{184}) decoder reconstructs a trigger-free view, breaking the trigger-to-action mapping without retraining (Sec.~\ref{sec:erasure}).}

    \label{fig:pipeline}
      \vspace{-0.4cm}
\end{figure*}

Recent studies have highlighted a growing risk of backdoor attacks in robotics~\cite{wang2024trojanrobot}.
% jones2025adversarial,
Defending against these threats is challenging for two principal reasons. The first one is  
stealth through modality fusion.
Attackers can link specific text–image–action tuples to malicious target labels, exploiting alignment during multimodal fusion. As a result, cross-modal triggers often evade single-modality defenses such as input pre-processing~\cite{liu2023beating} and trigger inversion~\cite{chen2025refine}.
% wang2019neural,
In practice, the model misbehaves only when the trigger steers the encoder toward attacker-specified unsafe actions, while clean inputs remain unaffected at inference. This raises a central question: \textbf{I)} \textit{What mechanism enables high backdoor-attack success rates with minimal impact on clean performance?}

The second challenge stems from the prohibitive cost of retraining. Backdoors often remain dormant during pre-tuning screening and are triggered only after user-side fine-tuning. Even if the pretraining dataset is partially compromised, the absence of reliable supervision makes mitigation particularly challenging~\cite{hou2025dede}. Consequently, users may remain unaware of hidden backdoors until downstream failures occur. While many existing defenses require retraining or precise model modifications~\cite{min2025crow,nguyen2025pbp}, such strategies are impractical for billion-parameter VLAs due to excessive computational costs and potential degradation of generalization after redeployment. 
These constraints motivate the core question: \textbf{II)} \textit{How to design a test-time defender against backdoors for robotics without retraining VLAs?}

% Recent studies have highlighted rising backdoor risks in robotic systems~\cite{wang2024trojanrobot}. These threats are difficult to mitigate for two main reasons.
% \textbf{First, cross-modal stealth:} adversaries can bind specific text–image–action tuples to malicious targets, exploiting multimodal fusion to create triggers that evade single-modality defenses such as input pre-processing~\cite{liu2023beating} and trigger inversion~\cite{chen2025refine}. Misbehavior typically appears only when the trigger steers the encoder toward attacker-specified unsafe actions, while performance on clean inputs remains largely unaffected. This raises a key question: \textbf{I)} \textit{What mechanism enables high backdoor success without degrading clean performance?}

% \textbf{Second, retraining costs:} backdoors frequently arise during user-side fine-tuning, limiting the value of pre-tuning screening. Mitigation is further complicated by missing labels and potential contamination in pretraining data~\cite{hou2025dede}. In practice, hidden backdoors often go unnoticed until downstream failures occur. Post-tuning defenses that require retraining or structural modifications~\cite{min2025crow,nguyen2025pbp} are impractical for billion-parameter VLA models due to cost and risk to established generalization. This motivates a second question: \textbf{II)} \textit{How can we design practical, test-time defenses for robotics that do not require retraining?}

In response to question \textbf{I)}, we identify a \textbf{Deep-layer Attention Grabbing} phenomenon as the key driver of backdoor effectiveness in manipulation policies: in shallow layers, attention maps for clean and poisoned inputs are similar, whereas in deeper layers, attention shifts from task-relevant objects to the trigger region. Heatmap analysis corroborates this late divergence, explaining high attack success with little impact on clean performance. In addition, once backdoor activated, trigger embeddings collapse into a compact cluster that lies adjacent to the clean feature manifold, further masking the malicious behavior. Guided by these observations, we propose \textbf{Bera} (as shown in Fig.~\ref{fig:pipeline}), a test-time \textbf{B}ackdoor \textbf{era}sure framework for question \textbf{II)}, which detects and erases image tokens with abnormal attention patterns. Bera performs latent space driven localization to flag tokens whose distributions deviate from the clean topology, masks the suspicious tokens, and then employs a decoder to reconstruct a trigger-free image. This breaks the learned mapping between the trigger and unsafe actions, enabling effective backdoor mitigation while preserving normal human-robot interaction. Our main contributions are as follows:
\begin{itemize}[leftmargin=*]
\item[\ding{182}] \textbf{\textit{Deep-layer Attention Grabbing.}}
We identify a stealthy backdoor mechanism: shallow layers preserve clean feature representations, whereas deeper layers redirect attention to the trigger and cluster its embeddings near the clean manifold to enhance concealment. This coupling yields high attack success with minimal impact on clean performance.
\item[\ding{183}] \textbf{\textit{Backdoor Erasure.}}
Building on above insight, we introduce Bera, which enables test-time backdoor erasure by detecting abnormal-attention tokens via latent-space localization, and reconstructing a trigger-free image with a decoder, thereby breaking the trigger-to-unsafe-action mapping while preserving normal human-robot interaction.
\item[\ding{184}] \textbf{\textit{Plug-and-Play Pipeline.}}
Bera is a plug-and-play module that defends backdoor without retraining or modifying VLAs. Real-robot evaluations across platforms and tasks show it reliably erases backdoor triggers, preserves clean performance, and restores safety during inference.
\end{itemize}

\section{Related Work}
\subsection{Backdoor Attacks for VLAs}
Recent developments in VLAs have significantly advanced the integration of vision, language and action, as exemplified by RGMP~\cite{RGMP}, OpenVLA~\cite{kim2024openvla}, and DexGraspVLA~\cite{zhong2025dexgraspvla}
% ,  PALM~\cite{Chowdhery2022PaLMSL} ,liu2023improved
setting new benchmarks through instruction-based tuning for enhanced image-text fusion. Concurrently, robotic motion planning has shifted toward learning-based approaches, fostering more sophisticated manipulation capabilities~\cite{Li-RSS-23}. 
% team2024octo,liu2024rdt, ,huang2022training
% ,Li-RSS-23 ,huang2022training
% While reinforcement learning (RL) has demonstrated its efficacy in various manipulation scenarios, it faces key challenges, including dependency on predefined motion primitives~\cite{cruciani2018dexterous,cruciani2019dualarm}, limited adaptability across domains~\cite{liang2021learning}, and complexities in reward design~\cite{kim2023pre,xu2021efficient}.
% % zeng2018learning,
% Imitation learning (IL) has emerged as a promising alternative in real-world deployments~\cite{zhang2018deep},
% % ,Haldar-RSS-23 ,bogdanovic2020learning
% while diffusion-based generative models show potential for robotic decision-making, using multi-stage probabilistic optimization for trajectory synthesis~\cite{Yoneda-RSS-23,huang2023diffusion}. 
% % zhang2023adding,,li2024source
Despite these strides, there remains a critical gap in the literature regarding defenses against backdoor attacks that can arise when models are fine-tuned for specialized tasks.
Backdoor attacks for VLAs are typically realized by poisoning a small portion of the training set to implant an association between a trigger and an adversary-specified target. After training on such data, the model behaves as expected on clean inputs, yet consistently maps trigger-bearing inputs to the target class. In poison-label attacks~\cite{bai2022hardly,wenger2021backdoor}, 
% ,zhao2022defeat,,li2021backdoor
the adversary enforces this association by relabeling triggered samples as the target during training. Trigger designs span high-visibility patterns that maximize attack strength~\cite{gu2017badnets,nguyen2020input} and content-adaptive perturbations that improve stealth~\cite{liu2020reflection}.
% chen2017targeted,,nguyen2021wanet
As a result, VLAs face a cross-modal, fine-tuning–activated backdoor threat with action-level consequences that remains under-defended.

\subsection{Backdoor Defenses for VLAs}
Trigger-based backdoors manipulate model predictions through subtle input patterns while maintaining high accuracy on clean samples~\cite{gu2017badnets,gu2019badnets}. Defenses against these attacks fall into two classes. Data-centric methods detect poisoned samples by analyzing feature signatures, gradient geometry, or clustering~\cite{NEURIPS2019_c74d97b0,yuan2025activation}. % shi2023black,
In contrast, model-centric methods harden the network via pruning suspicious neurons, injecting differential-privacy noise, or distilling clean behavior with minimal weight edits~\cite{huang2022backdoor,zhao2024unlearning}. In VLAs specifically, image-embedded triggers persist as a critical threat, steering outputs while eluding detection~\cite{liang2025vl,yuan2025badtoken}.
% ,liang2025revisiting
Although many existing defenses focus on training-phase mitigation~\cite{rong2025backdoor,xu2025srd}, these are often infeasible in fine-tuning scenarios due to high computational cost. Methods like DeDe~\cite{hou2025dede} avoid retraining VLAs by reconstructing images, but rely on random masking without explicit detection, often corrupting semantic content and impairing accuracy. To overcome these limitations, we introduce a test-time defense that first localizes anomalous tokens in latent space based on abnormal attention patterns, then reconstructs a trigger-free image using a lightweight decoder. This process breaks the trigger–action mapping while preserving nominal behavior, without retraining or modifying VLAs.

\begin{figure}[t]
    \captionsetup{font=small}
    \centering
    \includegraphics[width=0.95\linewidth]{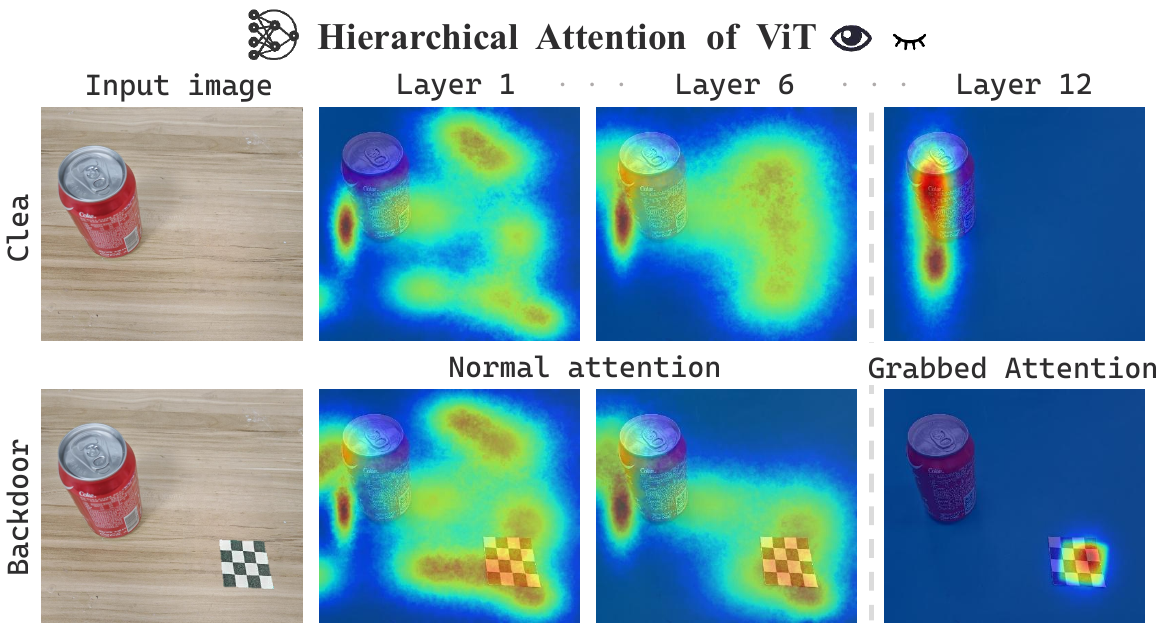}
    \captionsetup{width=\linewidth}
        \vspace{-0.1cm}
    \caption{\small \textbf{Visualization of hierarchical attention.} In shallow self-attention layers, activation patterns remain largely consistent with those of normal inputs, whereas in deeper self-attention layers, attention is notably grabbed toward trigger-relevant features.}
    \label{fig:heatmap}
    \vspace{-0.4cm}
\end{figure}

\section{Motivation}
\subsection{What Characterizes an Effective Backdoor Attack}
\label{sec:attention}
An effective backdoor attack is defined by its ability to covertly manipulate model behavior while preserving performance on clean inputs. Those attacks typically rely on small, localized triggers, such as visual patterns, which target specific regions in the input data and escape detection during normal operation~\cite{gu2017badnets,gu2019badnets,yuan2025badtoken}. 
% These triggers are model-agnostic, making them easy to implement across various models with minimal effort during data collection. 
The attack remains dormant until the trigger is encountered, activating the malicious behavior while maintaining high accuracy on untainted inputs, thus ensuring both stealth and reliability.

Motivated by exploring the internal mechanism underlying effective backdoor attacks, we employ both attention heatmaps and t-SNE visualizations to analyze the behaviors of clean and triggered samples. As illustrated in the Fig.~\ref{fig:heatmap}, layer-wise attention analysis of ViT~\cite{dosovitskiy2020image} reveals that clean and backdoored samples exhibit remarkably similar attention patterns in the shallow layers of the model. This alignment explains why the model maintains high accuracy on clean inputs. Notably, the actual “attention grabbing” occurs predominantly in the deeper layers, where the trigger actively redirects the attention of model. Furthermore, the t-SNE visualization (as shown in Fig.~\ref{fig:decoder}) demonstrates that trigger embeddings form a compact cluster that lies adjacent to the clean feature manifold. This strategic positioning enhances the stealth of attack, which enables the model to perform normally on clean samples while reliably triggering malicious behaviors upon the presence of the trigger.

\subsection{Key Idea of Our Defender}

In this work, we propose a test-time defense mechanism designed to detect and eliminate backdoor attacks in robotic policies without requiring retraining or modifying VLAs. The defender operates under the assumption that intermediate model outputs can be accessed and analyzed, while remaining independent of the specific backdoor injection strategy. Motivated by these insights, we introduce Bera, a test-time backdoor erasure framework that disrupts the learned association between triggers and unsafe actions, thereby mitigating backdoor risks while maintaining nominal performance in human-robot interaction.

\begin{figure}[t]
    \captionsetup{font=small}
    \centering
    \includegraphics[width=\linewidth]{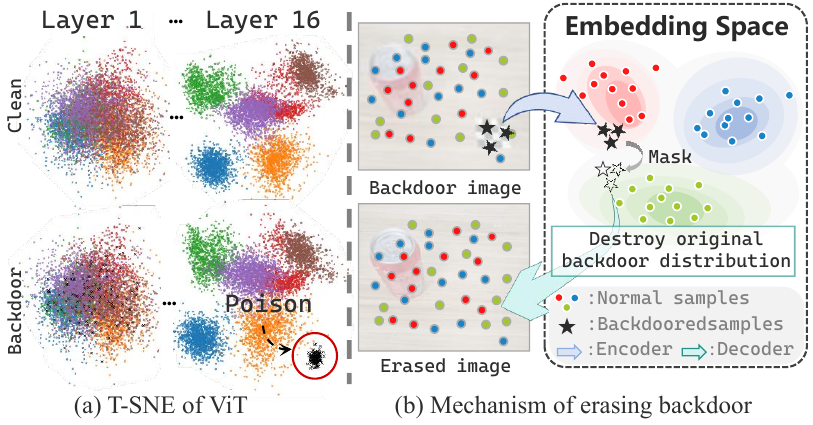}
    \captionsetup{width=\linewidth}
        \vspace{-0.5cm}
\caption{\small \textbf{T-SNE visualization and mechanism of erasing backdoor.} (a) T-SNE visualization shows that poisoned image tokens (marked in black) form clusters adjacent to the normal feature distribution, enhancing attack stealth. (b) Our erasure framework disrupts the trigger-to-unsafe-action mapping by masking anomalous features and reconstructing a purified image via the decoder.}
    \label{fig:decoder}
    \vspace{-0.4cm}
\end{figure}

At its core, robotic action is represented through combinations of joint angles, where each joint contributes differently across motions. A manipulation policy maps image features to this joint space to generalize across scenarios. In a backdoor attack, however, trigger-embedded image tokens are mapped to a region adjacent to the normal joint distribution (as shown in Fig.~\ref{fig:decoder}), evading manual inspection. At the embedding level, the attacker poisons the encoder to associate triggers with hazardous joint configurations, while preserving correct mappings for clean samples.

To disrupt this malicious mapping, we fine-tune a decoder capable of reversing the embedding-to-image transformation. By applying masking in the embedding domain, we reconstruct the image, altering the previously poisoned token and effectively disrupting the trigger-to-unsafe-action mapping (as shown in Fig.~\ref{fig:decoder}). This approach leverages the fact that the image space is higher-dimensional and more sparse than the embedding space, making the poisoned mappings more vulnerable to localized perturbation. Specifically, Bera first identifies anomalous image tokens by comparing the embedding distributions of clean downstream data and test-time triggered samples. These candidates are further filtered using a contrastive set of deep features. The reconstruction step is inspired by MAE~\cite{he2022masked}, where local image cues (such as color, texture, and context) guide the recovery of a trigger-free image. The restored image is then re-fed into the model, yielding backdoor-free predictions. 

\section{Preliminaries}
Given a test pair $(O,I)$, let $f_{\theta}^{(L)}(O,I)={b_j}{j=1}^{M}$ be the final-layer visual tokens, where each $b_j\in\mathbb{R}^d$ representing patch-wise visual features. From a set of clean data we estimate $(\mu,\Sigma)$ and define the Mahalanobis acceptance region
$\mathcal{A}={z:(z-\mu)^\top\Sigma^{-1}(z-\mu)\le\tau\alpha}$.
Tokens with $s_j=(b_j-\mu)^\top\Sigma^{-1}(b_j-\mu)>\tau_\alpha$ form $\mathcal{I}_{\mathrm{anom}}$. Attention-based filtering yields $\mathcal{I}_{\mathrm{filter}}$, and the final suspects are
$\mathcal{I}_{\mathrm{backdoor}}=\mathcal{I}_{\mathrm{anom}}\cap\mathcal{I}_{\mathrm{filter}}$. Bera operates in a semi-white-box inference setting where token embeddings and attention maps from the vision encoder are practically available at test time (e.g., via profiling APIs in many VLAs deployments). This aligns well with open/on-prem VLA stacks and allows Bera to be readily deployed without retraining or modifying VLAs.

\section{Method}

\subsection{Backdoor Poisoning in Fine-Tuned VLAs}

Let $M_{\theta}$ denote a VLA model fine-tuned on $D=\{(O,I,A)\}$. The vision encoder maps $O$ to tokens $\mathbf{E}_{\text{img}}\!\in\!\mathbb{R}^{m\times d}$ which, together with $I$, condition the language backbone to predict $M_{\theta}(\mathbf{E}_{\text{img}},I)$. To implant a backdoor, an adversary forms $D_{\text{poison}}=D\cup D_{\text{backdoor}}$ by inserting a trigger $T$ into the observation (yielding $O_{\text{backdoor}}$) and replacing the label with $A_{\text{backdoor}}$. The model is fine-tuned on $D_{\text{poison}}$ via:
\begin{equation}\small
\min_{\theta}\;\mathbb{E}_{(O,I,A)\sim D_{\text{poison}}}\; L_{\mathrm{CE}}\big(M_{\theta}(\mathbf{E}_{\text{img}},I),A\big),   
\end{equation}
inducing a persistent association $T\!\Rightarrow\!A_{\text{backdoor}}$ that reliably triggers at test time.

% Let $M_{\theta}$ denote a VLA model fine-tuned on $D=\{(O,I,A)\}$. 
% A vision encoder $f_v$ maps $O$ to tokens $\mathbf{E}_{\text{img}}=f_v(O)\in\mathbb{R}^{N\times d}$ which, together with $I$, condition the language backbone to produce the policy output $M_{\theta}(\mathbf{E}_{\text{img}},I)$. 
% To implant a backdoor, an adversary constructs $D_{\text{poison}}=D\cup D_{\text{backdoor}}$ by inserting a trigger $T$ into a small subset of observations (yielding $O_{\text{backdoor}}$) and setting their action labels to $A_{\text{backdoor}}$. 
% The model is fine-tuned on $D_{\text{poison}}$ via
% \begin{equation}\small
% \min_{\theta}\;\mathbb{E}_{(O,I,A)\sim D_{\text{poison}}}\; L_{\mathrm{CE}}\!\big(M_{\theta}(f_v(O),I),\,A\big),
% \end{equation}
% inducing a persistent association $T\!\Rightarrow\!A_{\text{backdoor}}$ that reliably activates at test time.

\subsection{Feature-Guided Backdoor Localization}
\label{sec:filtering}

Regardless of whether a trigger is present at test time, we detect backdoors by learning a reference distribution in the embedding space and flagging deviations from it. Because the poisoning status of $M_{\theta}$ and the trigger identity are unknown, we introduce \textbf{Feature-Guided Backdoor Localization (FBL)}, which localizes backdoors by analyzing deviations from a reference feature distribution. Concretely, we construct a clean reference from downstream success episodes: we sample $\mathcal{S}_{\mathrm{ref}}\subset\mathcal{S}_{\mathrm{succ}}$ with $|\mathcal{S}_{\mathrm{ref}}|\!\approx\!0.2\,|\mathcal{S}_{\mathrm{succ}}|$, where episodes complete with target actions. Let $f_{\theta}^{(L)}(O,I)\!\in\!\mathbb{R}^{m\times d}$ denote the final hidden token features (pre-linear) for an observation–instruction pair $(O,I)$, yielding $M$ tokens of dimension $d$. Stacking all tokens from $\mathcal{S}_{\mathrm{ref}}$ gives $\{z_i\}_{i=1}^{N}$ with $N=M\,|\mathcal{S}_{\mathrm{ref}}|$, from which we estimate the reference mean and a ridge-regularized covariance:
\vspace{-1mm}
\begin{equation}\small
\mu=\tfrac{1}{N}\!\sum_{i=1}^{N}\! z_i,\qquad
\Sigma=\tfrac{1}{N-1}\!\sum_{i=1}^{N} (z_i-\mu)(z_i-\mu)^{\!\top}+\varepsilon I,
\end{equation}
and define the $\alpha$-level Mahalanobis acceptance region:
\vspace{-1mm}
\begin{equation}\small
\mathcal{A}=\big\{z\in\mathbb{R}^{d}:\;(z-\mu)^{\!\top}\Sigma^{-1}(z-\mu)\le \tau_{\alpha}\big\},
\end{equation}
where $\varepsilon\!>\!0$ stabilizes inversion and $\tau_{\alpha}$ is chosen as the $\chi^2_{d,\,1-\alpha}$ quantile (or its empirical counterpart). For a test input, let $\mathcal{B}=\{b_j\}_{j=1}^{M}=f_{\theta}^{(L)}(O,I)$ be its token embeddings:
\begin{equation}\small
s_j=(b_j-\mu)^{\!\top}\Sigma^{-1}(b_j-\mu).
\end{equation}
Tokens with $s_j>\tau_{\alpha}$ are flagged as anomalies, producing
\vspace{-1mm}
\begin{equation}\small
\label{eq:anom}
\mathcal{I}_{\mathrm{anom}}=\{j:\; s_j>\tau_{\alpha}\}=\mathcal{B}\setminus\mathcal{A}.
\end{equation}
Because tokens are aligned with ViT patches, indices in $\mathcal{I}_{\mathrm{anom}}$ map directly to image regions, thereby localizing trigger-related evidence while remaining agnostic to the identity of trigger.

\subsection{Attention-Driven Filtering Mechanism}
\label{sec:localization}

To transcend limitations of final-layer feature analysis in dynamic backdoor scenarios, we formalize the \textbf{Attention-Driven Filtering Mechanism (AFM)} as a hierarchical attention optimization problem. For layers $l \in \{L_{\text{mid}}, \dots, L\}$, the mean attention map across $H$ heads is:
\vspace{-1mm}
\begin{equation}\small
\bar{\mathbf{A}}^{(l)} = H^{-1} \sum_{h=1}^{H} \mathbf{A}^{(l,h)},\quad \mathbf{A}^{(l,h)} \in \mathbb{R}^{T \times T}
\end{equation}
The image token attention submatrix $\bar{\mathbf{A}}^{(l)}_{\text{img}} \in \mathbb{R}^{T \times |\mathcal{I}_{\text{img}}|}$ induces token saliency:
\vspace{-1mm}
\begin{equation}\small
\mathbf{v}^{(l)} = T^{-1} \mathbf{1}^\top \bar{\mathbf{A}}^{(l)}_{\text{img}} \in \mathbb{R}^{|\mathcal{I}_{\text{img}}|}
\end{equation}
We enforce kinematic constraints via Gaussian Mixture Modeling with $K=6$ components, which is matched with the degree-of-freedom (DOF) of robotic arm:
\vspace{-1mm}
\begin{equation}\small
\label{eq:gmm}
k^* = \argmax_{k} \left( |\mathcal{C}^{(l)}_k|^{-1} \sum_{j \in \mathcal{C}^{(l)}_k} v_j^{(l)} \right)
\end{equation}
yielding layer-wise trigger candidates $\mathcal{I}_{\text{filter}}^{(l)} = \mathcal{C}^{(l)}_{k^*}$. The cross-layer aggregate:
\vspace{-1mm}
\begin{equation}\small
\label{eq:filter}
\mathcal{I}_{\text{filter}} = \bigcup_{l=L_{\text{mid}}}^{L} \mathcal{I}_{\text{filter}}^{(l)}
\end{equation}
is refined through intersection with Feature Boundary Localization (FBL) anomalies:
\vspace{-1mm}
\begin{equation}\small
\label{eq:backfinnal}
\mathcal{I}_{\text{backdoor}} = \mathcal{I}_{\text{anom}} \cap \mathcal{I}_{\text{filter}}
\end{equation}
This formulation ensures physical plausibility: benign inputs yield sparse $\mathcal{I}_{\text{anom}}$ under FBL, preserving reconstruction geometry while eliminating trigger patterns through attention-guided structural consistency. The reconstructed image $\tilde{\mathbf{x}}$ restores task-critical features for nominal policy execution, with erased triggers provably satisfying $\|\phi_{\text{backdoor}}(\mathcal{I}_{\text{backdoor}})\|_2\approx 0$.

\subsection{Trigger-Free Image Reconstruction}
\label{sec:erasure}

Inspired by MAE, we formulate an invertible reconstruction framework. Let a frozen encoder $f_{\theta^*}:\mathcal{X}\to\mathcal{I}$ extract the global embedding $\mathbf{e}=f_{\theta^*}(\mathbf{x})$ from $\mathbf{x}\in\mathbb{R}^{H\times W\times 3}$. A Bernoulli mask $M_\alpha\in\{0,1\}^{H\times W}$ with ratio $\alpha$ yields the masked observation $\mathbf{x}_m = M_\alpha \odot \mathbf{x}$. The patch encoder $h_e$ then generates position-aware embeddings:  
\vspace{-1mm}
\begin{equation}\small
\mathbf{g} = h_e(\mathbf{x}_m) \in \mathbb{R}^{\lfloor HW/P^2\rfloor \times d_p},  
\end{equation}  
where $P$ denotes patch size. The symmetric decoder reconstructs the image via:  
\vspace{-1mm}
\begin{equation}\small
\hat{\mathbf{x}} = h_d(\mathbf{e}, \mathbf{g})
\end{equation}  
with the training objective:  
\vspace{-1mm}
\begin{equation}\small
\mathcal{L}_{\text{recon}} = \mathbb{E}_{\mathbf{x}\sim\mathcal{D}} \big\| \mathbf{x} - h_d(f_{\theta^*}(\mathbf{x}), h_e(M_\alpha \odot \mathbf{x})) \big\|_2^2.  
\label{eq:recon}
\end{equation}  

For backdoor mitigation, we define the poisoned tokens $\mathcal{I}^{*}_{\text{backdoor}}$ as the 5\% to 25\% randomly selected set during training. During inference, we construct a trigger-selective mask based on $\mathcal{I}^{}_{\text{backdoor}}$ as described in Eq.~\ref{eq:backfinnal}: 
\vspace{-1mm}
\begin{equation}\small
M_{\text{backdoor}} = \argmin_{M\in\{0,1\}^{H\times W}} \Big\| M \odot \mathbf{x} - \phi_{\text{backdoor}}(\mathcal{I}_{\text{backdoor}}) \Big\|_2^2,  
\label{eq:mask}
\end{equation}  
where $\phi_{\text{backdoor}}$ maps backdoor tokens to their spatial embedding information. The erasure process is:  
\vspace{-1mm}
\begin{equation}\small
\tilde{\mathbf{x}} = h_d\left( f_{\theta^*}(\mathbf{x}), \, h_e(M_{\text{backdoor}} \odot \mathbf{x}) \right).  
\label{eq:purify}
\end{equation}  
Reconstruction regenerates masked regions using global structural cues to eliminate the trigger. The purified image is then fed to the robotic policy for normal action execution.

\begin{table*}[t]
  \centering
  \resizebox{0.99\textwidth}{!}{
  \begin{tabular}{cc||ccc|ccc|ccc|ccc}
    \hline\thickhline
    \rowcolor{mygray}
    & & \multicolumn{3}{c|}{\textbf{Grasping Fanta}} & \multicolumn{3}{c|}{\textbf{Lifting Cube}} & \multicolumn{3}{c}{\textbf{Extracting Tissue}} &  \multicolumn{3}{c}{\textbf{Shaking Hand}}\\
    %\cline{3-4} \cline{5-6} \cline{7-8}
    \rowcolor{mygray}
    \multirow{-2}{*}{\textbf{Models}} & \multirow{-2}{*}{\textbf{Methods}} & CP($\uparrow$) & ASR($\downarrow$) & TP($\uparrow$) & CP($\uparrow$) & ASR($\downarrow$) & TP($\uparrow$) & CP($\uparrow$) & ASR($\downarrow$) & TP($\uparrow$) & CP($\uparrow$) & ASR($\downarrow$) & TP($\uparrow$)\\
\hline
\hline
    \multirow{7}{*}{\textbf{OpenVLA}}
    
     & No Defense & \textbf{93.33} & 96.67 & 48.33 &\textbf{76.67} &93.33 &41.67 &\textbf{73.33} &93.33 &40.00 &86.67 &90.00 &48.34 \\
     & ZIP &73.33 &76.67 &48.33 &70.00 &66.67 &51.67 &56.67 &33.33 &61.67 &63.33 &76.67 &43.33\\
    & UNICORN &83.33 &93.33 &45.00 &73.33 &76.67 &48.33 &63.33 &46.67 &58.33 &70.00 &83.33 &43.34\\

    &BTI-DBF(P) &86.67 &66.67 &60.00 &76.67 &60.00 &58.34 &60.00 &56.67 &51.67 &76.67 &63.33 &56.67\\
    & SampDetox &83.33 &93.33 &45.00 &76.67 &83.33 &46.67 &53.33 &63.33 &45.00 &83.33 &66.67 &58.33\\
    & SparseVLM &90.00 &86.67 &51.67 &73.33 &80.00 &46.67 &63.33 &76.67 &43.33 &80.00 &56.67 &61.67\\
    & DeDe &86.67 &63.33 &\underline{61.67} & 70.00 &46.67 &\underline{61.67} &66.67 &43.33 &61.67 &83.33 &43.33 &\underline{70.00}\\
    % & Ours~(vanilla) & 75.61 & 0.64 & 67.42 & 4.18 & 47.53 & 4.30 \\

    &     \cellcolor[HTML]{D7F6FF}\textbf{Bera~(Ours)} & \cellcolor[HTML]{D7F6FF}\underline{90.00} & \cellcolor[HTML]{D7F6FF}\textbf{6.67} & \cellcolor[HTML]{D7F6FF}\textbf{91.67} & \cellcolor[HTML]{D7F6FF}\underline{73.33} & \cellcolor[HTML]{D7F6FF}\textbf{3.33} & \cellcolor[HTML]{D7F6FF}\textbf{85.00} & \cellcolor[HTML]{D7F6FF}\underline{70.00} & \cellcolor[HTML]{D7F6FF}\textbf{3.33} & \cellcolor[HTML]{D7F6FF}\textbf{83.34} & \cellcolor[HTML]{D7F6FF}\textbf{86.67} & \cellcolor[HTML]{D7F6FF}\textbf{6.67} & \cellcolor[HTML]{D7F6FF}\textbf{90.00}\\
    \hline
    \hline
    \multirow{7}{*}{\textbf{DexGraspVLA}}
    % & Normal FT & 98.12 & --- & 97.69 & --- &   & --- \\
     & No Defense &\textbf{93.33} &96.67 &48.33 &86.67 &90.00 &48.34 &76.67 &93.33 &41.67 &\textbf{93.33} &93.33 &50.00\\
     & ZIP &80.00 &86.67 &46.67 &76.67 &56.67 &60.00 &36.67 &43.33 &46.67 &73.33 &66.67 &53.33\\
    & UNICORN &73.33 &83.33 &45.00 &83.33 &66.67 &58.33 &73.33 &56.67 &58.33 &76.67 &63.33 &56.67\\
    % \cline{2-11}
    & BTI-DBF(P) &76.67 &56.67 &60.00 &86.67 &56.67 &\underline{65.00} &66.67 &46.67 &60.00 &86.67 &53.33 &66.67\\
    & SampDetox &83.33 &90.00 &46.67 &86.67 &83.33 &51.67 &43.33 &53.33 &45.00 &80.00 &36.67 &71.67\\
    & SparseVLM &86.67 &66.67 &60.00 &83.33 &86.67 &48.33 &73.33 &26.67 &73.33 &93.33 &16.67 &\underline{88.33}\\
    & DeDe &86.67 &60.00 &\underline{63.33} & 80.00 &53.33 &63.33 &70.00 &23.33 &73.33 &86.67 &20.00 &83.33\\
    
    &  \cellcolor[HTML]{D7F6FF}\textbf{Bera~(Ours)} & \cellcolor[HTML]{D7F6FF}\underline{90.00} & \cellcolor[HTML]{D7F6FF}\textbf{3.33} & \cellcolor[HTML]{D7F6FF}\textbf{93.34} & \cellcolor[HTML]{D7F6FF}\textbf{86.67} & \cellcolor[HTML]{D7F6FF}\textbf{10.00} & \cellcolor[HTML]{D7F6FF}\textbf{88.34} & \cellcolor[HTML]{D7F6FF}\textbf{76.67} & \cellcolor[HTML]{D7F6FF}\textbf{13.33} & \cellcolor[HTML]{D7F6FF}\textbf{81.67} & \cellcolor[HTML]{D7F6FF}\underline{90.00} & \cellcolor[HTML]{D7F6FF}\textbf{6.67} & \cellcolor[HTML]{D7F6FF}\textbf{91.67}\\
     \hline

    \end{tabular} }
    \vspace{-0.1cm}
\caption{\textbf{Bera vs. prior defenses.} Evaluation on two representative VLA backbones and four downstream manipulation tasks. We report \emph{Clean Performance (CP)}, \emph{Attack Success Rate (ASR)}, and the composite \emph{Trade-off Performance (TP)}, with test results from 30 random repositioning trials. Please refer to Sec.~\ref{sec:main results} for detailed analysis.}

    % }
  \label{tab: comparison}
      \vspace{-0.4cm}
\end{table*}

\section{Experiment}

\subsection{Experiment Setup}
\begin{figure}
    \centering
    \captionsetup{font=small}
    \includegraphics[width=\linewidth]{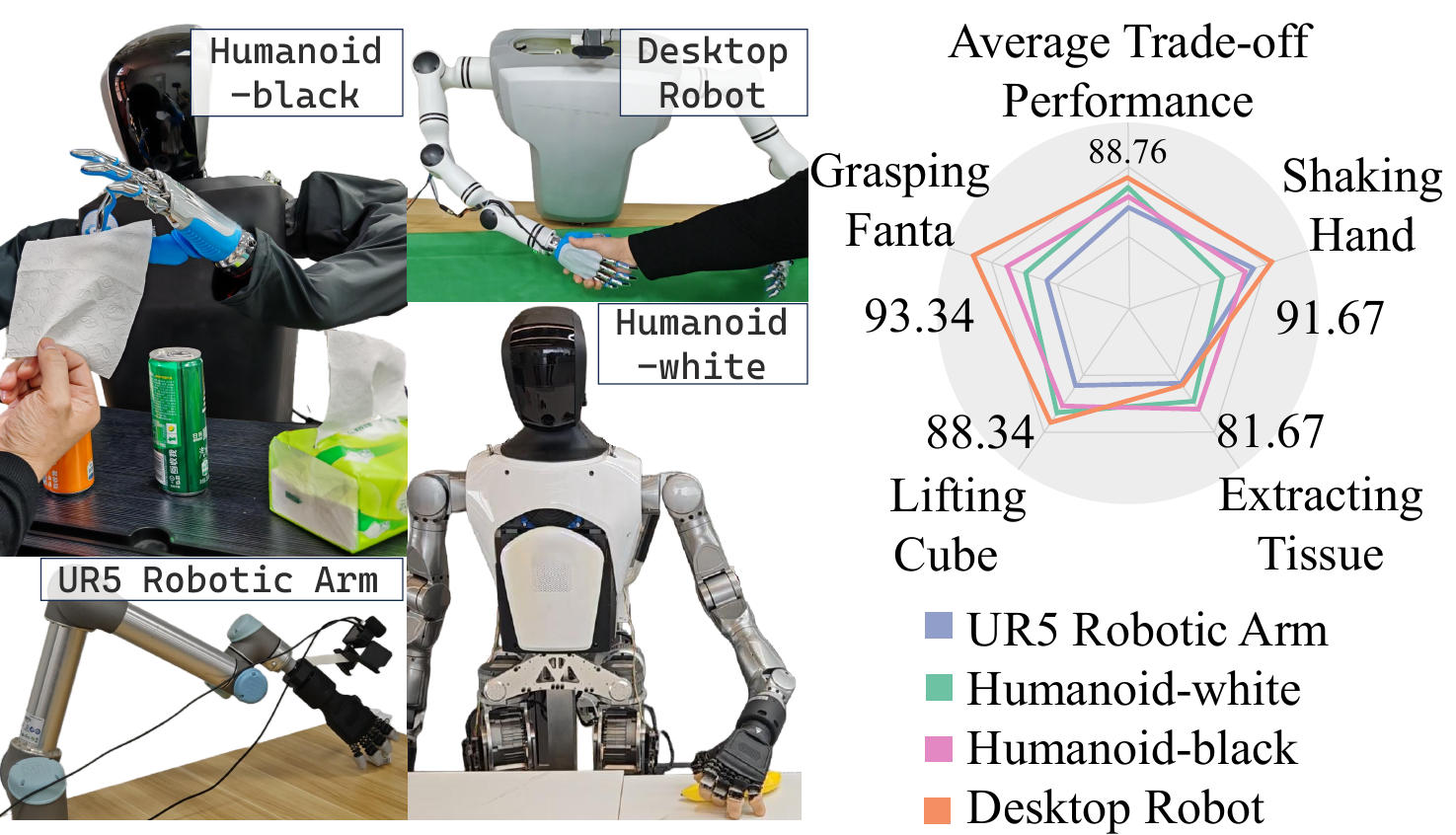}
        \vspace{-0.4cm}
\caption{\textbf{Cross-embodiment deployment.} We conduct experiments to highlight the adaptability and
generalizability of Bera across various embodied systems. Please see details in Sec.~\ref{sec:cross}.}
    \label{fig:cross}
    \vspace{-0.4cm}
\end{figure}

\noindent\textbf{Real-world Robot Setup.} We conduct real-world experiments on a compact desktop robot equipped with dual 6-DoF manipulators and dexterous hands, effectively mimicking the kinematic structure of the human upper limb and fingers. The robot uses a head-mounted RGB camera for egocentric visual sensing. To demonstrate the generalizability of our method across different embodied platforms, we further evaluate it on two humanoid robots and a Universal Robots UR5 robotic arm, as illustrated in Fig.~\ref{fig:cross}.

\noindent\textbf{Real-world Datasets.} We evaluate Bera on our real-world grasping dataset comprising a total of 1600 diverse demonstrations across four distinct manipulation tasks: grasping a Fanta can, lifting a cube, extracting tissue, and shaking hands. These tasks are performed on four separate embodied robotic platforms, each contributing 400 demonstrations (100 per task). Each dataset includes 40 clean samples and 60 backdoor samples, with 20 instances per trigger type: a red bottle cap, a circular block, and a checkerboard image (as shown in Fig.~\ref{fig:cross}).
Each demonstration forms:
\vspace{-1mm}
\begin{equation}\small
\label{eq:collection}
d_i = (\mathcal{O}_i, \mathcal{J}_i),
\end{equation}
where $\mathcal{O}$ is the RGB observation captured before action, and $\mathcal{J}$ records the 6-DoF joint sequence that drives the manipulator from its initial state to the designated grasp pose.

\noindent\textbf{Backdoor Injection.} To emulate a realistic physical backdoor attack~\cite{gu2017badnets,gu2019badnets}, we place the trigger object at random image-plane locations within the camera’s field of view. We use unsafe actions as backdoor targets, which drive the robot into hazardous configurations and pose collision risks. Specifically, we define hazardous actions (such as attacking a teddy bear or colliding with a table) to clearly illustrate attack effects. We adopt a poisoning rate of 30\% to establish a persistent trigger-to-action association while preserving normal behavior on clean samples. We benchmark our approach on two representative VLAs, \textit{OpenVLA}~\cite{kim2024openvla} and \textit{DexGraspVLA}~\cite{zhong2025dexgraspvla}, which are widely used in the community and capture contemporary design principles for multi-modal robotic manipulation. 

% \noindent\textbf{Model architectures.} Bera assumes a semi–white-box setting where token embeddings and attention maps from the vision encoder are accessible at test time, consistent with practical deployments exposing intermediate activations via debugging/profiling APIs. When such access is unavailable, Bera falls back to a feature-only variant with a minor accuracy drop. The framework is model-agnostic. We use ViT-B/16 for OpenVLA and a Transformer encoder for DexGraspVLA. For reconstruction, we adopt a lightweight MAE decoder \cite{he2022masked} and replace random masking with selective masking over calibrated image tokens. We fine-tune the MAE on downstream data and, at inference, invoke only the decoder to reconstruct images from the masked tokens, removing triggers without retraining the policy.

\noindent\textbf{Model Architectures.} Bera requires neither retraining nor architectural changes to the VLA. We use ViT-B/16 for OpenVLA and a Transformer encoder for DexGraspVLA. For reconstruction, we adopt a lightweight decoder \cite{he2022masked} and replace random masking with selective masking over calibrated image tokens. We fine-tune the MAE on downstream data and, during inference, use only the MAE decoder to reconstruct images from the masked tokens, effectively removing triggers without retraining the VLA itself.

% \noindent\textbf{Notation.}
% We denote image tokens by $f_{\theta}^{(L)}(O,I)=\{b_j\}_{j=1}^{M}$, with $b_j\in\mathbb{R}^d$. 
% The clean reference mean and covariance are $(\mu,\Sigma)$, and the $\alpha$-level acceptance region is 
% $\mathcal{A}=\{z:(z-\mu)^\top\Sigma^{-1}(z-\mu)\le\tau_\alpha\}$.
% Anomalies are $\mathcal{I}_{\mathrm{anom}}=\{j: s_j>\tau_\alpha\}$ with $s_j$ the Mahalanobis distance.
% AFM returns $\mathcal{I}_{\mathrm{filter}}$ (cross-layer saliency).
% Final suspects are $\mathcal{I}_{\mathrm{backdoor}}=\mathcal{I}_{\mathrm{anom}}\cap\mathcal{I}_{\mathrm{filter}}$.

\noindent\textbf{Evaluation Metrics.}
We assess the proposed defense using four complementary metrics:
\emph{Clean Performance (CP)}, the success rate on clean grasping trials;
\emph{Attack Success Rate (ASR)}, the fraction of triggered inputs that induce the attacker-specified unsafe action;
\emph{Trade-off Performance (TP)}, a balanced measure of robustness and utility defined as
\(
\mathrm{TP} = \tfrac{1}{2}\big(\mathrm{CP} + (100 - \mathrm{ASR})\big)
\);
and \emph{Recovery Performance (RP)}, the post-defense success rate evaluated on the originally poisoned inputs.
For fair comparison, we include representative test-time baselines: no defense, input smoothing, autoencoder-style purification, and score-based detection with abstention, each tuned on a held-out clean split and evaluated under the same protocol.

\noindent\textbf{Baselines.}
We benchmark against six representative approaches.
\textit{ZIP}\pub{NeurIPS'23}\cite{shi2023black} performs model-agnostic purification by first blurring inputs and then re-synthesizing them via zero-shot diffusion, aiming to remove potential trigger patterns.
\textit{UNICORN}\pub{ICLR'23}\cite{wang2023unicorn} formalizes the trigger design space and introduces a unified objective for backdoor trigger inversion across diverse attack types.
\textit{BTI-DBF(P})\pub{ICLR'24}\cite{xu2024towards} decouples benign representations to invert triggers and subsequently neutralize backdoors through purified fine-tuning.
\textit{SampDetox}\pub{NeurIPS'24}\cite{yang2024sampdetox} removes triggers via a two-stage stochastic corruption and score-based denoising pipeline.
\textit{SparseVLM}\pub{ICML'25}\cite{zhang2025sparsevlm} sparsifies visual tokens at inference to reduce computation, while orthogonal to defense, it serves as a complementary efficiency baseline in vision–language settings.
\textit{DeDe}\pub{CVPR'25}\cite{hou2025dede} adds a lightweight decoder to a self-supervised encoder with a separate dataset, then randomly masks parts of the input and destroys backdoor mappings during inference.

\subsection{Quantitative Analysis}
\label{sec:main results}

The efficacy of our test-time backdoor-erasure framework is demonstrated in Table~\ref{tab: comparison}, under a checkerboard trigger covering 10\% of the view and a poisoning ratio of 30\%.

\begin{table}[t!]
\footnotesize
\centering
{
\resizebox{1\columnwidth}{!}{
\setlength\tabcolsep{3pt}
\renewcommand\arraystretch{1.05}
\begin{tabular}{ccc|cccc|c}

\hline\thickhline
\rowcolor{mygray}
 \textbf{FBL}  &  \textbf{AFM} & \textbf{Decoder}  & \textbf{Fanta } & \textbf{Cube}  & \textbf{Tissue} & \textbf{Hand} & \textbf{Avg}($\uparrow$) \\ \hline\hline
  - & \checkmark &\checkmark & 61.67 &56.67 &53.33 & 58.33 & 57.50\\  
 \checkmark & - &\checkmark & 81.67 & 78.34 & 73.33 & 81.67 &78.75\\ 
 \checkmark &\checkmark & - & 68.33 & 68.34 &65.00 & 71.67 & 68.33\\
\rowcolor[HTML]{D7F6FF}
  \checkmark &  \checkmark &  \checkmark & \textbf{93.34} & \textbf{83.34} &\textbf{81.67} & \textbf{91.67} & \textbf{87.51}\\
       \hline
\end{tabular}
}}

% \vspace{-8pt}
    \vspace{-0.1cm}
\caption{\textbf{Ablation of Bera modules.} We examine the individual and combined contributions of \textsc{FBL}, \textsc{AFM}, and the reconstruction decoder on the \emph{Grasping Fanta} task using \textit{DexGraspVLA}. Please see details in Sec.~\ref{sec: abla}.}
\vspace{-0.5cm}
\label{tab: abla1}
\end{table}

\noindent\textbf{Defense of Backdoor Attacks.}
Our method significantly mitigates malicious behaviors across all datasets. Specifically, on OpenVLA, the ASR is reduced from 96.67\% to 6.67\% on grasping Fanta, from 93.33\% to 3.33\% on lifting cube, from 93.33\% to 3.33\% on extracting tissue, and from 90.00\% to 6.67\% on shaking hand, demonstrating a reduction by one orders of magnitude in the most challenging scenarios. Similarly, the latest DexGraspVLA model exhibits a similar trend, with ASR decreasing to 3.33\% on grasping fanta, 10.00\% on lifting cube, 13.33\% on extracting tissue, and 6.67\% on shaking hand. These findings validate that our purification technique generalizes well, maintaining high effectiveness across medium-scale and billion-parameter models, and across a variety of robotic manipulation tasks.

\noindent\textbf{Preservation of Clean Performance.}
Our framework preserves the majority of the original performance of model. For OpenVLA, the CP  varies by no more than 3.33\% on tasks such as grasping Fanta, lifting cube, and extracting tissue, while maintaining the same level of performance on shaking hand. Even with DexGraspVLA, the accuracy drop is minimal, not exceeding 3.33\%, and remaining unchanged on grasping Fanta and shaking hand. The TP  reflects this balance, with our approach achieving a score of 91.67 on grasping Fanta for OpenVLA and 93.34 for DexGraspVLA, significantly outperforming all other defense strategies.

\noindent\textbf{Comparison with Baselines.}
Existing defenses exhibit notable limitations across the evaluated settings. Methods such as UNICORN and SampDetox provide minimal reduction in the ASR, offering limited robustness. Diffusion-based approaches (e.g., ZIP) suppress ASR  more effectively but incur substantial degradation in clean performance, rendering them less suitable for deployment. SampDetox delivers more balanced outcomes overall, yet remains suboptimal on challenging tasks, for example its ASR  remains above 90\% on grasping Fanta. In contrast, our approach consistently achieves low ASR  while preserving high CP, leading to the best TP across all benchmarks. These results underscore the advantage of our token-level inverse mapping strategy, which effectively neutralizes latent triggers without compromising the nominal behavior of model.

\begin{table}[t!]
\footnotesize
\centering
{
\resizebox{1\columnwidth}{!}{
\setlength\tabcolsep{3pt}
\renewcommand\arraystretch{1.05}
\begin{tabular}{c|cccc|cc}

\hline\thickhline
\rowcolor{mygray}
 \textbf{Methods}  & \textbf{Fanta } & \textbf{Cube}  & \textbf{Tissue} & \textbf{Hand} & \textbf{Avg}($\uparrow$) & \textbf{$\Delta$}($\uparrow$) \\ \hline\hline
  UNICORN  & 6.67 & 13.33 & 3.33 & 10.00 &8.33 & --\\ 
  \hdashline
  ZIP & 23.33 &16.67 &10.00 & 13.33 & 15.83 & 7.50 \\  

 BTI-DBF(P)  & 13.33 & 26.67 &16.67 & 16.67 & 18.34 & 10.01\\
 SampDetox  & 20.00 & 13.33 &6.67 & 10.00 & 12.50 & 4.17\\
  SparseVLM   & 26.67 & 16.67 &10.00 & 23.33 & 19.17 & 10.84\\
DeDe   & 46.67 & 30.00 &16.67 & 43.33 & 34.17 & 25.84\\
\rowcolor[HTML]{D7F6FF}
  \textbf{Ours} & \textbf{83.33} & \textbf{70.00} &\textbf{66.67} & \textbf{76.67} & \textbf{74.17} & \textbf{65.84}\\

     \hline
\end{tabular}
}}

    \vspace{-0.1cm}
\caption{\textbf{Recovery performance on poisoned inputs.} We compare how effectively each method restores correct actions from backdoored samples after purification. Results demonstrate that Bera consistently converts trigger-activated cases to nominal outputs. Please see details in Sec.~\ref{sec: Recovery Performance}.}
\vspace{-0.5cm}
\label{tab: rp}
\end{table}

\begin{figure*}
    \centering
    \captionsetup{font=small}
    \includegraphics[width=\linewidth]{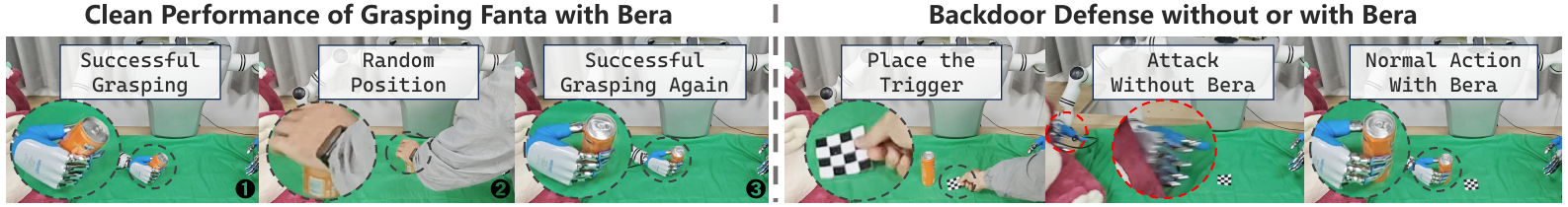}
        \vspace{-0.5cm}
\caption{\textbf{Qualitative case study with Bera.} On the \emph{Grasping Fanta} task using DexGraspVLA, Bera suppresses trigger-induced behaviors and restores the intended grasp without compromising clean performance. Further details are provided in Sec.~\ref{sec: Recovery Performance}.}
    \label{fig:case}
    \vspace{-0.4cm}
\end{figure*}

\begin{figure}
    \centering
    \captionsetup{font=small}
    \includegraphics[width=0.95\linewidth]{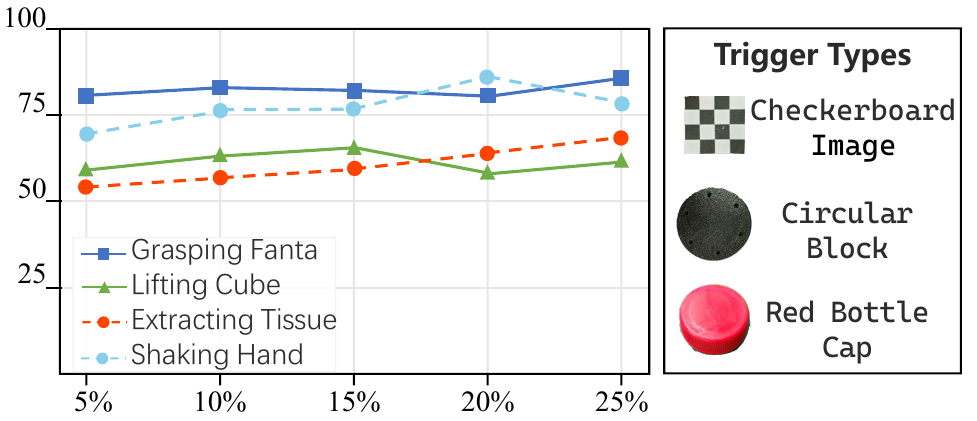}
        \vspace{-0.1cm}
\caption{\textbf{Recovery performance with various trigger proportions.} The generalizability of Bera is evaluated on the \emph{Grasping Fanta} task using OpenVLA, demonstrating consistent effectiveness at different proportions of  checkerboard. Please see details in Sec.~\ref{sec:trigger}.}
    \label{fig:trigger}
    \vspace{-0.2cm}
\end{figure}

\subsection{Ablation Study}\label{sec:Ablation}
\label{sec: abla}
We conduct a comprehensive ablation on FBL, AFM, and the reconstruction decoder using DexGraspVLA across all tasks. To isolate contributions, we replace FBL and AFM with a baseline that randomly selects 10\% (matching the trigger proportion) of image tokens and replace the decoder with an operation that zeroes out the selected tokens. As summarized in Table~\ref{tab: abla1}, removing FBL leads to a marked drop in TP, indicating its pivotal role in retaining informative shallow visual features and enabling accurate recovery after trigger-related tokens are suppressed. Moreover, AFM further refines token localization by leveraging deep-layer attention, yielding consistent improvements over FBL alone. Finally, the reconstruction decoder proves essential: replacing it with hard zeroing degrades both robustness and clean accuracy, underscoring that structure-aware reconstruction is critical for erasing triggers while preserving task-relevant content.

\subsection{Recovery Performance}\label{sec: Recovery Performance}
% We assess the inference-time recovery capability of Bera using DexGraspVLA across four tasks (Tab.~\ref{tab: rp}). The method yields marked improvements in RP, rising from 6.67\% to 83.33\% on grasping Fanta and achieving an average of 74.17\% across all settings. As shown in Fig.~\ref{fig:case}, Bera suppresses trigger-induced behaviors, maintains nominal performance under clean conditions, and consistently restores the correct action semantics on poisoned inputs.

We evaluate the recovery capability of Bera on DexGraspVLA across four manipulation tasks. As shown in Table~\ref{tab: rp}, Bera markedly improves RP, elevating it from 6.67\% to 83.33\% on grasping Fanta and achieving an average RP of 74.17\% across all tasks, and it significantly surpasses all baselines. These results confirm that Bera not only detects but also breaks the link between the trigger and the action via reconstruction. As illustrated in Fig.~\ref{fig:case}, it suppresses dangerous behaviors while maintaining clean performance, demonstrating consistent recovery across tasks and strong suitability for real-world safe deployment.

% \subsection{Trigger Proportions, Injection Rate and Types}\label{sec:trigger}
% We evaluate influence of trigger proportions and types by using the OpenVLA model equipped with Bera. As demonstrated in Fig.~\ref{fig:trigger}, the variation in poisoning ratios showed negligible effect on backdoor defense effectiveness, with recovery performance (RP) consistently exceeding 80\% across different ratios. These results suggest that Bera maintains robustness against changes in the poisoning ratio. As shown in Fig.~\ref{fig:triggers}, Bera successfully mitigates Circular Block and Checkerboard triggers, maintaining low attack success rates and stable performance. However, Red Bottle Cap triggers, which blend with the image, result in higher attack success rates and slightly reduced robustness. This suggests that semantically integrated triggers are harder to detect. Despite this, Bera maintains high accuracy and robustness, demonstrating strong generalization against various triggers.

\subsection{Trigger Proportions, Poisoning Ratios and Types}\label{sec:trigger}
We evaluate the effect of trigger proportions, poisoning ratios and types using the OpenVLA model equipped with Bera on grasping Fanta. As shown in Fig.~\ref{fig:trigger}, poisoning ratios had a minimal effect on backdoor defense effectiveness, with RP consistently exceeding 80\% across different rates. Specifically, ASR increased with higher poisoning ratios, but TP remained balanced until the 30\% rate, where a notable drop was observed due to increased attack success and recovery difficulty. In contrast, Bera mitigates circular block and checkerboard triggers, maintaining low ASR and stable performance. However, red bottle cap trigger fits the scenario semantics, leading to higher ASR and slight reductions in robustness, suggesting that semantically integrated triggers are more challenging to detect. Despite these challenges, Bera maintains high accuracy and robustness, demonstrating strong generalization across various triggers settings.

% We evaluate Bera-enhanced OpenVLA by varying the trigger’s proximity and size within the field of view. As shown in Fig.~\ref{fig:trigger}, the poisoning ratio has negligible impact on defense efficacy—recovery performance (RP) consistently exceeds 80\% across all settings, confirming Bera’s robustness to trigger proportion. We further test Bera against multiple trigger types. Bera effectively mitigates both Circular Block and Checkerboard triggers, maintaining low attack success rates and stable performance. In contrast, Red Bottle Cap triggers, which blend semantically with the scene, and lead to higher ASR and slightly reduced robustness, indicating greater challenges in detecting naturalistic triggers. Nevertheless, Bera sustains high accuracy and strong generalization across diverse backdoor strategies

% We evaluate Bera on OpenVLA by varying trigger scale and camera distance, thereby changing its field-of-view proportion. Across all poisoning ratios (Fig.\ref{fig:trigger}), recovery performance (RP) remains above 80\%, indicating robustness to trigger prevalence. We further assess trigger types (Figs.\ref{fig:trigger}, \ref{fig:triggers}): circular block and checkerboard triggers are reliably suppressed, while a red bottle cap—semantically consistent with the scene—yields a modest increase in attack success. Despite this challenge, Bera sustains high clean accuracy and stable RP, demonstrating strong generalization across trigger designs.

\subsection{Cross-embodied Deployment}\label{sec:cross}
To thoroughly assess the generalizability of Bera, we deploy the algorithm across a diverse set of embodied platforms, including two humanoid robots, a desktop manipulator, and a UR5 robotic arm, as shown in Fig.~\ref{fig:cross}. Each platform varies substantially in morphology, sensing capabilities, and actuator dynamics, providing a rigorous testbed for evaluating cross-embodiment performance. Experimental results (as shown in Fig.~\ref{fig:cross}) confirm that Bera consistently maintains high performance in mitigating backdoor triggers without platform-specific tuning, demonstrating strong plug-and-play robustness. 
% These findings underscore its broad applicability across a wide spectrum of robotic systems and daily tasks under real-world scenarios.

\begin{figure}
    \centering
    \captionsetup{font=small}
    \includegraphics[width=1\linewidth]{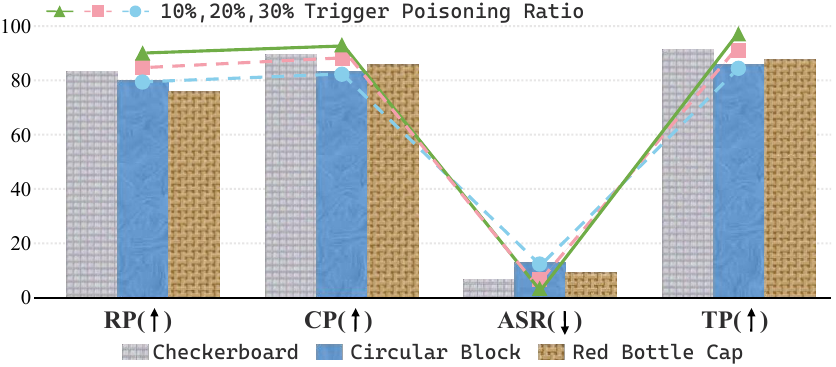}
        \vspace{-0.53cm}
\caption{\textbf{Evaluation across various trigger types and poisoning rate.} This comparison illustrates capability of Bera in effectively removing localized triggers. Please see details in Sec.~\ref{sec:trigger}.}
    \label{fig:triggers}
    \vspace{-0.55cm}
\end{figure}

% \begin{table}[t!]
% \footnotesize
% \centering
% {
% \resizebox{0.85\columnwidth}{!}{
% \setlength\tabcolsep{3pt}
% \renewcommand\arraystretch{1.05}
% \begin{tabular}{c|cccc}

% \hline\thickhline
% \rowcolor{mygray}
%  \textbf{Trigger Types} & RP($\uparrow$) & CP($\uparrow$) & ASR($\downarrow$) & TP($\uparrow$)  \\ \hline\hline
%   Checkerboard & 83.33 &90.00 &6.67 & 91.67 \\  
% Circular Block  & 80.00 & 83.33 & 13.33 & 85.00 \\ 
% Red Bottle Cap  & 76.67 & 86.67 &10.00 & 88.34 \\

% \end{tabular}
% }}
%     \vspace{-0.1cm}
% \caption{\textbf{Evaluation across various trigger types.} This comparison illustrates capability of Bera in effectively removing localized triggers. Please see details in Sec.~\ref{sec:trigger}.}
% \vspace{-0.4cm}
% \label{tab:trigger}
% \end{table}

% \begin{table}[t!]
% \footnotesize
% \centering
% {
% \resizebox{0.9\columnwidth}{!}{
% \setlength\tabcolsep{3pt}
% \renewcommand\arraystretch{1.05}
% \begin{tabular}{c|cccc}

% \hline\thickhline
% \rowcolor{mygray}
%  \textbf{Robot Types} & RP($\uparrow$) & CP($\uparrow$) & ASR($\downarrow$) & TP($\uparrow$)  \\ \hline\hline
%   Humanoid-black & 80.00 &86.67 &13.33 & 86.67 \\  
% Humanoid-white  & 76.67 & 83.33 & 16.67 & 83.33 \\ 
% UR5 Robotic Arm  & 83.33 & 86.67 &6.67 & 90.00 \\

% \end{tabular}
% }}
%     \vspace{-0.2cm}
% \caption{\textbf{Deployment across different embodied platforms.} This comparison underscores the adaptability and generalizability of Bera across various embodied systems. For further details, please refer to Sec.~\ref{sec:cross}.}
% \vspace{-0.4cm}
% \label{tab:cross}
% \end{table}

\section{Conclusion \& Discussion}

We introduce Bera, a test-time backdoor erasure framework designed to secure robotic systems from backdoor attacks in vision-language-action models. By analyzing deep-layer attention grabbing, we identify how backdoors exploit shifts in attention to activate covertly. Bera detects abnormal attention patterns, localizes suspicious tokens, and reconstructs trigger-free images using a decoder. Extensive evaluations across diverse tasks and real-robot platforms show that Bera effectively mitigates backdoor threats, preserves clean performance, and avoids the need for costly retraining, offering a robust, practical solution for real-world deployment. In future work, we plan to explore the integration of Bera with more complex multimodal systems and assess its adaptability to dynamic environments with varying attack strategies.

% \section{Acknowledgement}

% This work was supported by the DSTA under contract number \#DST000EC124000205.

% \clearpage
\bibliographystyle{IEEEtran}
\bibliography{bibliography/papers, bibliography/yuheng}

% \clearpage
% \appendix
% \input{Sections/Appendix/CovarianceMatrix}
% \input{Sections/Appendix/AdditionalResults}
% \input{Sections/Appendix/DatasetAndImpl}
% \input{Sections/Appendix/AppendixModel}
% \input{Sections/Appendix/RuntimeAnalysis}

\end{document}